\documentclass[letterpaper, 10 pt, conference]{ieeeconf}

\IEEEoverridecommandlockouts                              

\usepackage{bbding}
\usepackage{graphicx}%
\usepackage{multirow}%
\usepackage{amsmath,amssymb,amsfonts}%
\usepackage{subcaption}
\usepackage{mathrsfs}%
\usepackage{pifont}
\usepackage[table,xcdraw]{xcolor}
\title{
3DWG: 3D Weakly Supervised Visual Grounding via Category and Instance-Level Alignment
}

\author{Xiaoqi Li$^{1}$, Jiaming Liu$^{1}$, Nuowei Han$^{1}$, Liang Heng$^{1}$, Yandong Guo$^{2}$, Hao Dong$^{1}$, Yang Liu$^{3*}$
\thanks{$^{1}$School of CS, Peking University; $^{2}$AI2Robotic; $^{3}$Wangxuan Institute of Computer Technology, Peking University; $*$ corresponding author}
}

\begin{document}

\maketitle
\thispagestyle{empty}
\pagestyle{empty}

\begin{abstract}
The 3D weakly-supervised visual grounding task aims to localize oriented 3D boxes in point clouds based on natural language descriptions without requiring annotations to guide model learning.
This setting presents two primary challenges: category-level ambiguity and instance-level complexity.
Category-level ambiguity arises from representing objects of fine-grained categories in a highly sparse point cloud format, making category distinction challenging.
Instance-level complexity stems from multiple instances of the same category coexisting in a scene, leading to distractions during grounding.
To address these challenges, we propose a novel weakly-supervised grounding approach that explicitly differentiates between categories and instances.
In the category-level branch, we utilize extensive category knowledge from a pre-trained external detector to align object proposal features with sentence-level category features, thereby enhancing category awareness.
In the instance-level branch, we utilize spatial relationship descriptions from language queries to refine object proposal features, ensuring clear differentiation among objects.
These designs enable our model to accurately identify target-category objects while distinguishing instances within the same category. 
Compared to previous methods, our approach achieves state-of-the-art performance on three widely used benchmarks: Nr3D, Sr3D, and ScanRef.
\end{abstract}


\section{INTRODUCTION}\label{sec1}
\begin{figure*}[t]
    \centering
    \begin{minipage}{.35\textwidth}
        \centering
        \includegraphics[width=0.9\linewidth, height=0.15\textheight]{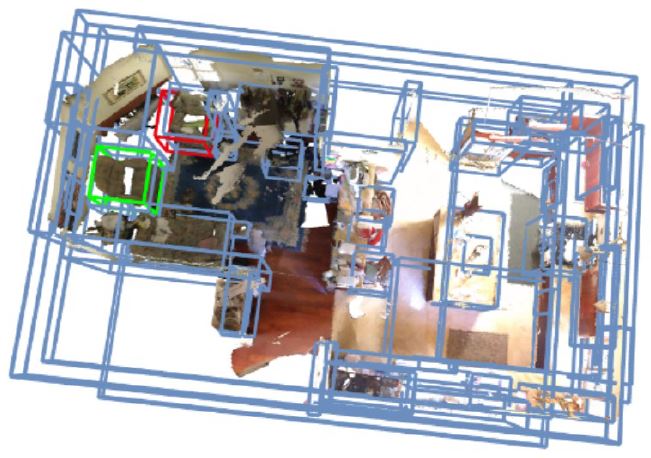}
        \subcaption{``find the armchair that is next to the table, which is also facing the window." }
    \end{minipage}%
    \begin{minipage}{.3\textwidth}
        \centering
        \includegraphics[width=0.9\linewidth, height=0.15\textheight]{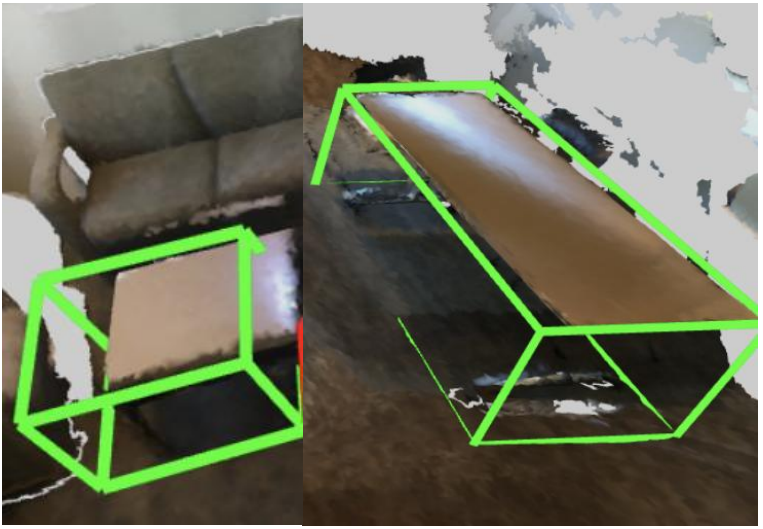}
        \subcaption{``end table" v.s. ``table" }
    \end{minipage}%
    \begin{minipage}{0.3\textwidth}
        \centering
        \includegraphics[width=0.9\linewidth, height=0.15\textheight]{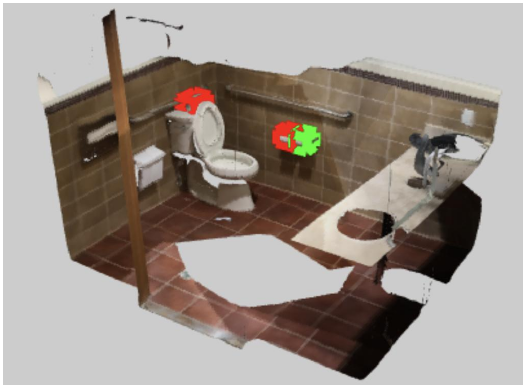}
        \subcaption{``Choose the toilet paper on the wall, closest to the sinks."}
        \label{fig:prob1_6_1}
    \end{minipage}
     \caption{\textbf{Challenges in 3D weakly visual grounding.} (a) and (b) show the category ambiguity. (c) demonstrates instance complexity. The {\color{green}green} box and {\color{red}red} boxes denote target instance and distracting instances respectively.}
     \vspace{-0.5cm}
    \label{fig:intro}
\end{figure*}
The 3D visual grounding task aims to localize oriented 3D boxes in a 3D scene based on natural language descriptions ~\cite{yang2021sat,yuan2021instancerefer,he2021transrefer3d,huang2022multi,roh2022languagerefer,guo2023viewrefer,zhao20213dvg,yang2024llm,yang2025lidar,pan2024renderocc}. 
This task has significant application potential in autonomous agents ~\cite{maes1993modeling,liu2024bevuda,liu2023vida,liu2024continual} and embodied AI~\cite{long2024discuss,wang2023find,li2024manipllm,liu2024robomamba,jia2024lift3d,xiong2024aic,huang2024manipvqa} and has achieved remarkable progress in recent years.
However, the fully-supervised setting requires extensive manual annotations, leading to high labor costs and limiting scalability and practical deployment in real-world applications. 
As a result, weakly supervised grounding has gained increasing attention ~\cite{wang2023distilling}.
In this setting, only the point cloud scene and its corresponding sentence descriptions are available during training, with no additional annotations provided.

The primary challenges of 3D weakly supervised grounding stem from its limitations at both the category and instance levels.
We begin with \textbf{category-level ambiguity}.
As illustrated in Fig.\ref{fig:intro}(a), 3D indoor scenes contain a diverse range of objects spanning multiple categories\cite{chen2020scanrefer,achlioptas2020referit3d}, such as ``armchair,” making accurate category identification challenging, especially when visually similar categories, such as ``chair,” coexist.
Furthermore, as shown in Fig.~\ref{fig:intro}(b), 3D point clouds are inherently sparse and often lack sufficient semantic information. This limitation is particularly pronounced in fine-grained indoor environments, where minimal semantic cues make it difficult to distinguish between similar categories. For instance, differentiating an ``end table” from a ``table” is challenging due to their highly similar appearances and shapes.
Regarding \textbf{instance-level complexity},
as depicted in Fig.~\ref{fig:intro}(c), 3D indoor scenes often contain multiple instances of the same category with nearly identical appearances. For example, a scene may include three identical toilet paper rolls, making it difficult to precisely distinguish and identify each individual instance.

To address these challenges, we propose a unified framework that tackles weakly supervised grounding at both the category and instance levels.
In the \textbf{category-level branch}, our goal is to enhance the model’s ability to distinguish categories, ensuring that the selected objects belong to the intended target category.
To achieve this, we first leverage category knowledge from a powerful pre-trained model to guide the alignment of object proposal features with the target category described in the sentence, ensuring precise category-level matching.
Furthermore, to mitigate category ambiguity caused by sparse point clouds and complex indoor environments, we introduce an explicit learning mechanism to differentiate easily confused categories. 
Specifically, we replace the target category phrase in the language query with similar but distinct categories, encouraging the model to bring object feature representations closer to positive language queries while pushing them away from negative ones.
In \textbf{instance-level branch}, we aim to tackle the challenge of distinguishing multiple instances of the same category by leveraging their unique 3D spatial relationships, thereby enhancing the model’s ability to differentiate instances.
First, we align the features of all objects and phrases, then refine object selection by identifying the object with the highest similarity score to the corresponding noun phrase in the query. Next, we classify the spatial relationship of the selected object based on the spatial descriptions provided in the query.
By incorporating spatial relationship reasoning, our approach effectively differentiates instances within the same category, enabling precise localization and identification of the queried object.

During the training phase, our framework simultaneously learns to align cross-modal feature representations at both the category and instance levels, without annotation supervision.
In the inference phase, it selects the instance based on the similarity score and confidence score by matching visual and language feature.
To evaluate our method, we conduct extensive experiments on three widely used benchmarks, \emph{i.e.}, Nr3D, Sr3D~\cite{achlioptas2020referit3d}, and ScanRef~\cite{chen2020scanrefer}.
Our approach outperforms existing weakly supervised grounding methods, demonstrating superior effectiveness in 3D visual grounding.


\section{RELATED WORK}\label{sec2}
\subsection{3D Supervised Visual Grounding}

3D visual grounding aims to identify the specific instance referred to in a given query within a 3D scene, represented as RGB-XYZ point clouds.
Two pioneering works, ReferIt3D~\cite{achlioptas2020referit3d} and ScanRef~\cite{chen2020scanrefer}, introduce benchmark datasets and baseline methods for 3D visual grounding.
Both studies select 3D scans from the ScanNet~\cite{dai2017scannet} dataset and enrich them with language query pairs and corresponding annotations.
Furthermore, an increasing number of studies~\cite{he2021transrefer3d,yuan2021instancerefer,wang2021improving,chen2018knowledge,yeh2018unsupervised,yeh2017interpretable,liu2021relation,gupta2020contrastive,khan2022weakly,wang20233drp,zhang2024vision,yuan2024visual,wang2024g,qian2024multi,zhang2024towards,xu2024multi,shi2024aware,zhang2024multi} have improved 3D visual grounding performance by introducing two-stage grounding frameworks.
Specifically, SAT~\cite{yang2021sat} leverages 2D information to enhance 3D grounding.
More recently, 3D-SPS~\cite{luo20223d} proposes a single-stage grounding method that selects key points and progressively refines the object distinction.
Additionally, $D^3$-LQ~\cite{wang2024g} models intrinsic geometric features and explicitly generates fine-grained object queries for 3D visual grounding.

However, 3D scene captures large context space of numerous objects, which are each annotated in the form of three dimension box. These fully supervised methods require laborious manual annotation of 3D box annotations, target box coordinate, and target category, thus limiting their scalability and practicability. Therefore, in this paper, we aim to propose a novel framework for weakly supervised 3D grounding by aligning cross-modality feature representation in both category and instance levels.

\subsection{3D Weakly Supervised Visual Grounding }
The task of 3D weakly visual grounding is to figure out specific instances only given the scene and language query pair.
In a prior study DCFVG~\cite{wang2023distilling}, an approach is devised to tackle the challenge of 3D weakly-supervised grounding, involving a two-step procedure.
First, it compares the similarity between the features of objects and the features of the target category and selects k candidates with highest similarity score.
Next, it refines the initial selection to determine the final object from among these k candidates by evaluating each object's similarity with a concealed language query. The concealed language query masks noun phrases and adjectives, intending to promote the model's comprehension of the query.

However, the initial phase encounters significant challenges in filtering out objects belonging to the target category. Specifically, matching numerous sparse point cloud objects from fine-grained categories with the target category phrase is highly challenging.
In contrast, our approach extracts target category knowledge from a powerful pre-trained model and leverages it to guide the matching process. This ensures a reliable alignment between object proposal features and the sentence’s target category, enhancing category awareness.
Furthermore, DCFVG aligns the features of the selected object candidates with the language query, where critical noun phrases and adjectives are concealed to promote a deeper understanding of the query. However, this approach has a key limitation: the initial selection of object candidates may already contain errors and fail to align with the noun phrase in the query, leading to error accumulation in instance differentiation.
In contrast, our method aligns \textit{each} object and phrase individually and incorporates spatial relationships from the language description to distinguish the corresponding object, enabling the model to accurately differentiate between potential instances.
\section{METHOD}\label{sec3}

\begin{figure*}[ht!]
\includegraphics[width=0.90\textwidth]{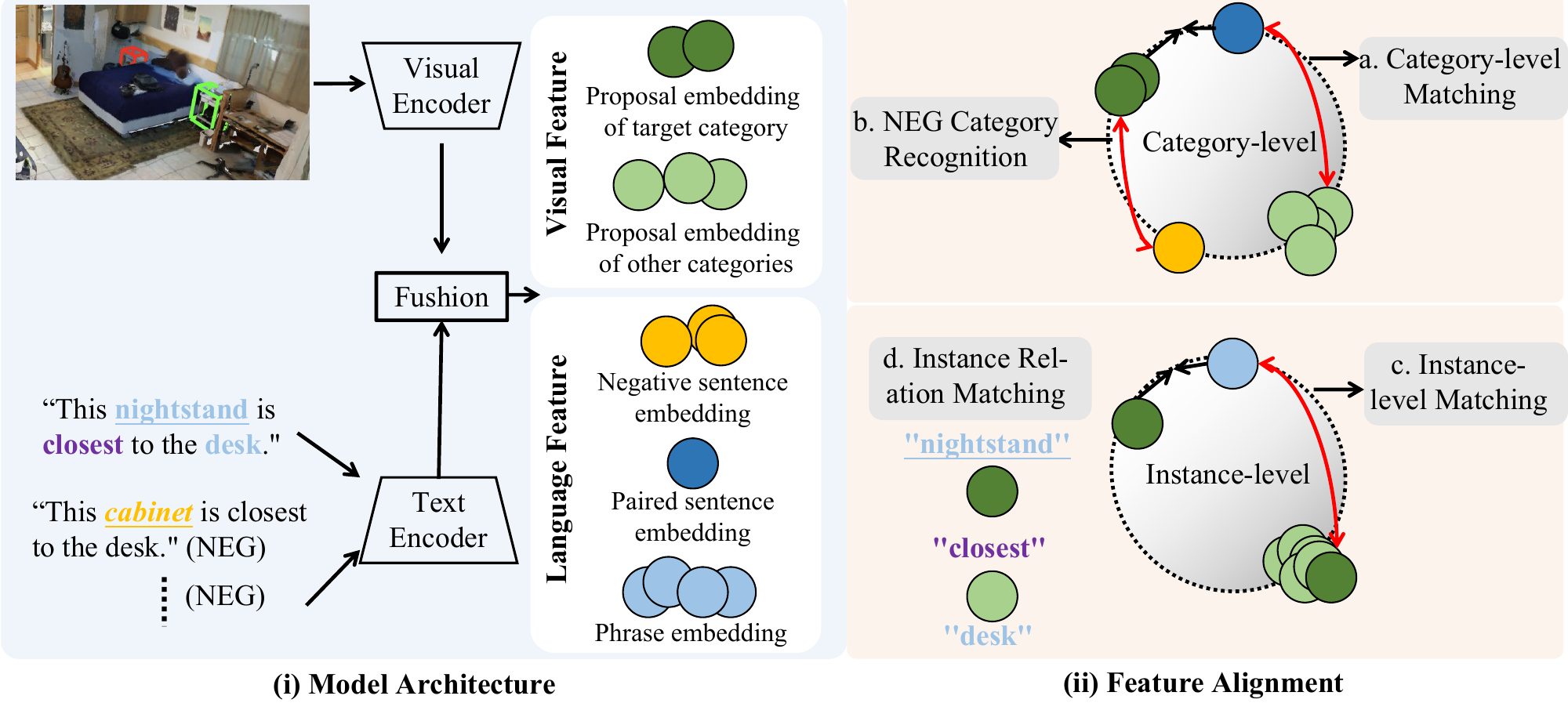}
\centering

\caption{\textbf{Illustration of our overall framework.} After obtaining features of both modalities through encoder and fusion modules, we adopt category-level and instance-level branched to realize category and instance identification. Both branches contribute jointly in inference under decision logic.}
\vspace{-0.6cm}
\label{fig:method}
\end{figure*}

\subsection{Model Architecture}
\label{sec:3.2}
In the weakly-supervised setting, given a 3D point cloud scene $\mathcal{S}$ of $N$ points ,  the goal is to locate target object that is described in the language query $\mathcal{Q}$. There are no bounding box annotations but only scene and language query pairs in the training stage. 

\textbf{Visual encoder.}
As shown in Fig.~\ref{fig:method}(i), given a 3D point cloud scene \(\mathcal{S}\) with \(N\) points, where each point has 3D coordinates and \(F\)-dimensional auxiliary features (e.g., RGB, normal vectors), we use a pre-trained detector to generate object candidate proposals. We then retain the top \(m\) proposals that exceed a predefined confidence threshold, which contrain proposals of target category and other categories.
Next, a visual encoder extracts proposal embeddings \( f_{\text{po}} \in \mathbb{R}^{m \times \mathcal{D}} \), where \( \mathcal{D} \) is the embedding dimension.  
\textbf{Text encoder.}
Given a query \( \mathcal{Q} \) with up to \(n\) words, we first convert words into tokens and then use a pre-trained BERT model~\cite{kenton2019bert} to generate token embeddings. The [CLS] token captures sentence-level category knowledge. During training, we fine-tune the BERT text encoder.  
\textbf{Fusion module.}
We use a stack of transformer layers to fuse embeddings from both modalities. The fused outputs include \textit{Proposal embedding} \( F_{\text{po}} \in \mathbb{R}^{m \times \mathcal{D}} \) and
\textit{Paired sentence embedding} \( F_{\text{se}} \in \mathbb{R}^{1 \times \mathcal{D}} \), which encodes the category semantics of the language query.
For \textit{phrase embeddings}, we first extract all noun phrases and their attributes using Scene Graph Parser~\cite{zellers2018neural}. From the language query, we extract up to \( n' \) noun phrases and obtain phrase embedding \( F_{\text{phr}} \in \mathbb{R}^{n' \times \mathcal{D}} \).  

\subsection{Feature Alignment}
As shown in Fig.~\ref{fig:method}(ii), we introduce strategies to align the extracted features at both the category and instance levels.

\subsubsection{Category-level Branch}
\label{sec:3.3}
Indoor 3D scenes contain various fine-grained categories that are sparsely represented and can be easily confused, making it challenging to accurately distinguish.
To address this issue, in the category-level branch, our goal is to enable the model distinguish different categories and become aware of the target category, ensuring to ground on object of correct category.

\textbf{a. Category-level Matching.} 
\label{sec:c1}
In this module, we focus on aligning the category-level semantic information of the language query sentence with the category semantics of the object visual proposal embeddings.
The goal is to ensure a reliable alignment and promote consistency between the feature representation of the sentence query and the object proposals that belong to target category, thus becoming target category aware.

Specifically, we map proposal embeddings \( F_{\text{po}} \in \mathbb{R}^{m \times \mathcal{D}} \) to category likelihoods \( P_{s} \in \mathbb{R}^{m \times c} \) using a classifier with an MLP structure, where \( c \) represents the total number of categories.  
We observe that the target category phrase is most commonly found in the subject token of the language query. To extract this phrase, we use Spacy~\cite{honnibal2015improved} and Scene Graph Parser~\cite{zellers2018neural} to analyze the language query and identify the target category phrase \( p_{d} \). 
As illustrated in Fig.~\ref{fig:prob1_6_1}, the word ``nightstand" is identified as the extracted target category.  
Among all the category likelihoods in \( P_{s} \), we retain only the likelihood corresponding to the target category as a supervision signal. We denote this as \( S_{se}' \in \mathbb{R}^{m \times 1} \), which represents the probability of each proposal belonging to the target category \( p_d \).  
Additionally, we compute the similarity scores between the sentence embedding $F_{se}$ and the proposal embeddings $F_{po}$ using the cosine similarity function.
The sentence embedding \( F_{se} \)extracted from the [CLS] token, encodes the target category features from the language query.
Finally, we use the category likelihood \( S_{se'} \) as a supervision signal to guide the similarity score \( S_{se} \) in cross-modality learning, optimized under the loss function \( \mathcal{L}_{se} \). This ensures that the target category representations from both modalities (language and vision) are pulled closer, allowing the grounding model to be target category aware and improving its ability to correctly identify objects of the intended category.


      
    

\textbf{b. Negative Category Recognition.} 
\label{sec:c2}
Sparse point cloud object is of fine-grained category and usually lacks sufficient semantic information, which will cause confusion in category distinction.
This motivates us to explicitly differentiate cross-modality representation of easy-confused categories.

Specifically, we generate $k$ negative language queries by replacing the target category phrase with other easy-confused noun phrases. 
We then calculate the cosine distance of target category phrase with other candidate noun phrases (all from the dataset) and adopt the top-$k$ similar phrases. 
Same as the process of extracting positive sentence embedding, we obtain the negative sentence embedding $F_{se}^{NEG}\in\mathbb{R}^{k\times\mathcal{D}}$, which represents the negative category semantic knowledge of the query. 
We evaluate the similarity scores between both positive and negative sentence embeddings and proposal embeddings by computing the cosine distance.  
Since direct query-proposal supervision is unavailable, we instead measure the similarity between the query and the scene, which can be supervised by the input pair.  
For a given query, we sum the top three highest query-proposal similarity scores to represent the query’s compatibility with the scene.  
Thus, we obtain the compatibility scores of \( k \) negative queries and one positive query with the scene, denoted as \( S_{PN}^{(\mathcal{Q}_{neg},\mathcal{S})} \in \mathbb{R}^{k} \) and \( S_{PN}^{(\mathcal{Q}_{pos},\mathcal{S})} \in \mathbb{R}^{1} \), respectively.  
These scores are then optimized using the InfoNCE loss~\cite{chen2020improved}, formulated as \( \mathcal{L}_{PN} \), which encourages the object proposal representations to be closer to positive sentence embeddings while pushing them away from negative sentence embeddings.  
Through training with our framework, the feature distribution of easily confused categories after the visual encoder exhibits disentanglement, demonstrating the effectiveness of our approach in differentiating categories.


\subsubsection{Instance-level Branch}
\label{sec:3.4}
In the indoor 3D scene, it is the common case that multiple instances of the same category and same appearance lie in one scene simultaneously, requiring more efforts of differentiating in instance-level. Meanwhile, we observe that instance can be specified from their contextual relation, especially in 3D scenario of abundant spatial relation information. This phenomenon promotes us to introduce the instance-level branch which compensates for instance uncertainty by leveraging their spatial relation.

\textbf{c. Instance-level Matching.}
\label{sec:f1}
Its goal is to realize a comprehensive instance-level alignment which can further facilitate specifying the target instance from distracting instances.
We first compute the similarity score \( s_{(x,y)} \) between \( n' \) phrases and \( m \) proposals by calculating their cosine distances, where \( x \in \{1,2,...,n^{'}\} \) and \( y \in \{1,2,...,m\} \).  
Since no fine-grained phrase-proposal supervision exists, we instead utilize query-scene pairs for supervision.  
Specifically, for each scene-query pair \(\mathbb{S}=\{\mathcal{S}_{1},...,\mathcal{S}_{b}\}\) and \(\mathbb{Q}=\{\mathcal{Q}_{1},...,\mathcal{Q}_{b}\}\) in a mini-batch of size \( b \), we measure all phrase-proposal similarity scores \( s_{(x,y)} \) within the pair.  
As shown in Eq.~\ref{eq:4}, for a given noun phrase, we select the maximum similarity score among all proposals.  
We then sum the maximum similarity scores of all phrases within the language query to represent the overall query-scene similarity \( S(\mathcal{Q}_i,\mathcal{S}_j) \).  
As shown in Eq.~\ref{eq:5}, this similarity score is then optimized using the InfoNCE loss \( \mathcal{L}_{phr} \), where \( p = 1 \) if \( \mathcal{Q}_i \) and \( \mathcal{S}_j \) are paired, and \( p = 0 \) otherwise.  
Here, \( \tau \) is the temperature scaling factor, and \( h(\mathcal{Q}_i,\mathcal{S}_j) \) is defined in Eq.~\ref{eq:6}.  
This approach encourages closer alignment between paired phrase-proposal representations while pushing apart unpaired instances, improving differentiation among instances.


\begin{equation}
\scriptsize
S(\mathcal{Q}_i,\mathcal{S}_j) = \sum_{x=1}^{n'} Max_{y\in\{1,2,..,m\}} (s_{(x,y)})
\label{eq:4}
\end{equation}

\begin{equation}
\scriptsize
\mathcal{L}_{phr.}(\mathcal{Q}_i,\mathcal{S}_j) = -\mathbb{E}_{\mathcal{S}_j\in\mathbb{S}}[p * log (h(\mathcal{Q}_i,\mathcal{S}_j))]\\
\label{eq:5}
\end{equation}

\begin{equation}
\scriptsize
h(\mathcal{Q}_i,\mathcal{S}_j) = \frac{exp(S(\mathcal{Q}_i,\mathcal{S}_j)/\tau)}{exp(S(\mathcal{Q}_i,\mathcal{S}_j)/\tau)+\sum_{\mathcal{S}_{j'}\in \mathbb{S}-\mathcal{S}_j}exp(S(\mathcal{Q}_{i},\mathcal{S}_{j'})/\tau)}
\label{eq:6}
\end{equation}

\textbf{d. Instance Relation Matching. } 
\label{sec:f2}
Since 3D scene is of three-dimensional space, there are rich spatial relationships between objects. Besides two-dimensional spatial relationships (such as left, right, up, down), there are also many three-dimensional spatial relationships (such as front and back).
This phenomenon promotes us to adopt 3D spatial relations in order to enrich instance spatial contextual information and realize instance identification.

Specifically, to extract spatial relation description in language query, we first build a relation library that consists of prevalent spatial relationships. We then judge whether there is relation phrase in language query that belongs to the relation library and lies between two noun phrases. If so, we will adopt it as the relation ground truth for these two noun phrases. Taking the input of Fig. \ref{fig:method} as an example, spatial relation ``closest'' is the relation ground truth for ``nightstand'' and  ``desk''. 
After establishing the ground-truth, we measure the similarity of two phrases embedding with all proposal embeddings and extract the most relevant proposal embedding $F_{po}^{i}$ and $F_{po}^{j}$ by calculating cosine distance. 
We then fuse the features of $F_{po}^{i}$ and $F_{po}^{j}$ to predict the spatial relationship of this two objects. 
It is then supervised by spatial relationship ground truth under classification cross-entropy loss $\mathcal{L}_{rel.}$, enabling the differentiation of instances through spatial context reasoning.

\subsection{Training Objectives and Inference}
\label{sec:3.5}

The two parallel category-level and instance-level branches are trained simultaneously and the overall loss $\mathcal{L}$ is: $\mathcal{L} = \lambda_1*\mathcal{L}_{se.}+\lambda_2*\mathcal{L}_{PN.} + \lambda_3*\mathcal{L}_{phr.}+\lambda_4*\mathcal{L}_{rel.}$, where $\lambda_1$, $\lambda_2$, $\lambda_3$, and $\lambda_4$ are hyper-parameters to balance each part.
During inference, the category-level branch calculates the cosine distance between all proposal embeddings and a sentence embedding denoted as $p_{c}\in\mathbb{R}^{m\times 1}$.
On the other hand, the instance-level branch computes the cosine distance for each possible proposal-phrase pair, resulting in a matrix $s_{f}\in\mathbb{R}^{m\times n'}$. Among these pair-wise similarities, we identify the highest similarity score for each proposal, represented as $p_{f}\in\mathbb{R}^{m\times 1}$.
To make a final prediction, we compare the maximum similarity score obtained from each branch and choose the branch with the higher maximum score. This selection reflects the branch's confidence in its prediction.

\section{EXPERIMENT}
\subsection{Implementation Details}

\begin{table*}[tb]
\caption{\textbf{Comparisons with baselines}. We compare our method with 3D-supervised, 2D weakly supervised, and 3D weakly supervised grounding methods. The highest score among all method types is underlined, while the highest score within the 3D weakly supervised setting is highlighted in bold.}
\vspace{-0.4cm}
	\begin{center}
	
	\small
	    \setlength{\tabcolsep}{1.3mm}{
		\begin{tabular}{c|c|ccc|ccc|cc}
			\hline
	\multirow{2}{*}{\textbf{}}&\multirow{2}{*}{\textbf{}}&\multicolumn{3}{c}{\textbf {Sr3D}}& 
			\multicolumn{3}{c}{\textbf {Nr3D}} &
			\multicolumn{2}{c}{\textbf {ScanRefer}}\\
			
		&Method&Acc@.25&Acc@.50&Acc&Acc@.25&Acc@.50&Acc& Acc@.25&Acc@.50 \\\hline
			
		Group 1: 3D	&ScanRefer\cite{chen2020scanrefer}        & - &-    &-   & - &- &-  &-  &22.4     \\
        Supervised Grounding&ReferIt3DNet\cite{achlioptas2020referit3d}        & 27.7 &-    &\underline{37.2}   &24.0  &-  &\underline{35.6}  & 26.4  &16.9     \\
		
			&SAT\cite{huang2021text}        & 44.5 &\underline{30.1}  &-     &-  &-  &-  & - &-       \\
			&InstanceRefer\cite{yuan2021instancerefer}  &31.5 &-    &-   &29.9  &- &-  & 40.2  &32.9    \\
			
			&BUTD-DETR\cite{jain2021looking}  & \underline{52.1}&-  &-   &\underline{43.3} &-   &- & \underline{52.2}  &\underline{39.8}     \\
			\hline
Group 2: 2D Weakly & A2G\cite{Datta_2019_ICCV}  &2.41 &3.15   &9.67    &3.29  &4.01 &12.38  &3.49 &4.28  \\
 Supervised Grounding&	Contr. Dist.\cite{gupta2020contrastive} 
  & 3.46 &5.70  &11.28     &3.82  &6.07 &15.26  &5.22 &7.67  \\
  &Contr. Learning\cite{wang2021improving} &3.51&4.76   &12.92    & 4.13  &6.13&15.63& 5.61 &7.23 \\\hline
		
	Group 3: 3D Weakly &	DCFVG\cite{wang2023distilling} &22.45   & 
    17.62 &-  & 22.45  & \underline{\textbf{17.62}} &-&27.37&21.96 \\
\rowcolor[HTML]{EFEFEF} Supervised Grounding&	Ours &\textbf{29.23}   & \textbf{19.16} &\textbf{36.5}  & \textbf{23.68}  & 15.30& \textbf{33.8}&\textbf{35.55}&\textbf{26.07} \\

			\hline
		\end{tabular}}
		
		\label{tab:main}
	\end{center}
	\vspace{-0.8cm}
\end{table*}
\textbf{Dataset.} We conduct experiments on three benchmarks Nr3D, Sr3D ~\cite{achlioptas2020referit3d}, and ScanRef ~\cite{chen2020scanrefer}. Natural Reference Nr3D contains 707 unique indoor scenes with 41503 natural language queries, where multiple distracting instances existed. Sr3D contains 83572 queries automatically generated based on a ``target''-``spatial relationship''-``anchor object'' template. ScanRef augments the 800 3D indoor scenes in the ScanNet~\cite{dai2017scannet} dataset with 51583 language queries. They all follow the official ScanNet ~\cite{dai2017scannet} train-test splits.

\textbf{Evaluation metrics.} We use the top-1 accuracy metric, which measures the percentage of times we can find the target box with an IoU higher than the threshold 0.25 and 0.5 IoU, denoted as Acc@.25 and Acc@.50 according to ~\cite{chen2020scanrefer}. In addition, following ~\cite{achlioptas2020referit3d}, when assuming the access to ground truth objects as the 3D proposals, we can convert the grounding task into a classification problem ~\cite{achlioptas2020referit3d}. The models then are evaluated by the accuracy Acc, i.e., whether it correctly selects the referred object among all proposals.

\textbf{Training details.} For a fair comparison, text encoder and fusion module all follow baseline SAT~\cite{yang2021sat}, while off-the-shelf visual encoder GroupFree~\cite{liu2021group} follows baseline BUTD-DETR ~\cite{jain2021looking} since SAT does not provide detector.
Specifically, we employ a text transformer with 3 layers and a fusion transformer with 4 layers.
The text transformer is initialized using the first three layers of BERT base~\cite{devlin2018bert}, while the fusion transformer is trained from scratch.  
We set the embedding dimension \( \mathcal{D} \) to 768 across all transformer layers.  
For language processing, we follow the maximum sentence length setting from ReferIt3D~\cite{achlioptas2020referit3d} and extract up to 16 noun phrases from each language query using Scene Graph Parser~\cite{zellers2018neural}.  
Additionally, we generate 25 negative queries corresponding to each query.  
In the overall objective function, the weighting factors are set as follows:  
\( \lambda_1 = 0.4 \), \( \lambda_2 = 0.005 \), \( \lambda_3 = 0.005 \), and \( \lambda_4 = 0.6 \).

\subsection{Compare with Baselines}

\textbf{Comparisons with 3D supervised approaches.}
Tab.~\ref{tab:main} Group 1 presents results across three benchmarks, comparing our method with fully supervised 3D visual grounding approaches that utilize annotations for 3D objects in the scene, the referred target object, and the target category.
Our method is under weakly-supervised training, with only language query and point cloud scene pairs. The weakly-supervised approach achieves competitive results compared with typical fully-supervised methods, demonstrating the benefit of the our training strategy.

\textbf{Comparisons with 2D weakly-supervised approaches.}
In Table \ref{tab:main} Group 2, we assess the effectiveness of a 2D weakly-supervised approach on three commonly used 3D grounding benchmarks. To ensure a fair comparison, we replace the 2D visual encoder with an off-the-shelf pretrained 3D encoder, GroupFree\cite{liu2021group}, which is identical to ours. Even with a pretrained detector, the 2D approaches exhibit poor performance, highlighting the need for methods specifically designed for 3D weakly supervised grounding.


\textbf{Comparisons with 3D weakly-supervised approaches.}
In Table \ref{tab:main} Group 3, we compare with a 3D weakly-supervised method using the same visual backbone and pretrained dataset to ensure a fair comparison. Our method surpasses DCFVG\cite{wang2023distilling} by a significant margin, demonstrating its effectiveness.
In the first stage, DCFVG uses a self-supervised method for matching, potentially leading to misalignment issues. In contrast, we utilize target category knowledge from a powerful off-the-shelf model, which ensures a relative reliable matching.
Moreover, DCFVG selects the top-k proposals based on the matching process and aligns the features of these selected object candidates with the language query. However, if the initially selected objects are incorrect, it can lead to error accumulation, further compromising the model’s accuracy.
Conversely, we align the feature of \textit{each} object with each phrase and incorporate spatial relation descriptions of noun phrases to distinguish objects, helping the model recognize different instances.

\subsection{Ablation Study and Analysis}
\label{sec:4.3}

\textbf{Effectiveness of each module.} 
As shown in Tab. \ref{tab:abla1}, we demonstrate the effectiveness of each module in our design. 
Starting from the first row ($c_1$ only), we adopt category-level matching to ensure target category identification, thus achieving relatively satisfying performance. When applying $c_2$ negative category recognition, we achieve a performance improvement since we boost category sensitivity by distinguishing target category and easy-confused categories, thus avoiding target on negative categories. Moreover, applying $i_1$ instance-level matching can further improve grounding ability since it formulates comprehensive instance-level discernment to extract the target instance from multiple instances of the same category. Finally, adding $i_2$ instance relation matching boosts the performance by adopting contextual relation constraints to differentiate each instance.

\begin{table}[h]
 \vspace{-0.2cm}
 \caption{\textbf{Ablation study.} We show the effectiveness of each designed component. } 
 \vspace{-0.4cm}
\small
 \begin{center}
  
  \small
    \setlength{\tabcolsep}{1.0mm}{
    \begin{tabular}{c|c|c|c|cccccc}
   \hline
    \multirow{2}{*}{\textbf{}}& \multirow{2}{*}{\textbf{}}& \multirow{2}{*}{\textbf{}}& \multirow{2}{*}{\textbf{}}&\multicolumn{2}{c}{\textbf {Sr3D}}& 
			\multicolumn{2}{c}{\textbf {Nr3D}} &
			\multicolumn{2}{c}{\textbf {ScanRefer}}\\
			$c_1$& $c_2$&$i_1$&$i_2	$& @.25&@.50&@.25&@.50& @.25&@.50\\\hline
			
   
   \Checkmark&-&-&-&22.1 &15.2&18.8&11.3&30.9&21.7\\
   \Checkmark&\Checkmark&-& -& 25.8&16.1&22.2&14.0 &34.1&24.3\\
   \Checkmark&\Checkmark&\Checkmark&-&  27.6&18.9 & 22.9&14.4&\textbf{35.7}&25.7\\
   \rowcolor[HTML]{EFEFEF} \Checkmark&\Checkmark&\Checkmark&\Checkmark & \textbf{29.2}&\textbf{19.2} & \textbf{23.7}&\textbf{15.3}&35.6&\textbf{26.1}  \\
     
    \hline
    \end{tabular}%

    \label{tab:abla1}%
    }
   
    	\end{center}
  \vspace{-0.6cm}
\end{table}

\begin{figure*}[h]
\includegraphics[width=15cm, height=7.5cm]{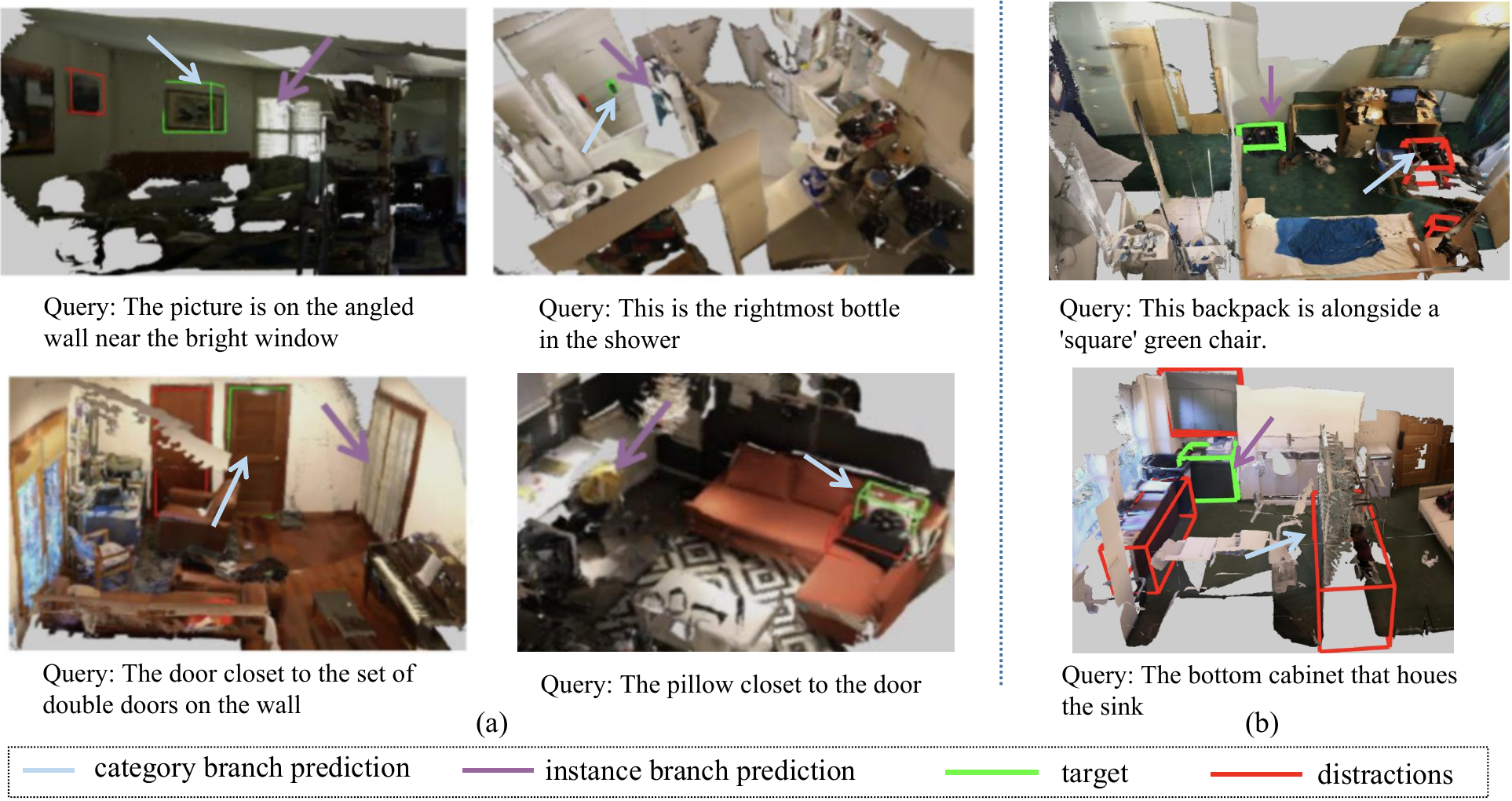}
\centering
\caption{\textbf{Visualization.} We present the predicted results at both category and instance levels, demonstrating their effectiveness }
\label{fig:vis_more}
\vspace{-0.7cm}
\end{figure*}

\textbf{Discussion on external detector. }
We aim to verify that the superior performance of our method is not obtained through the use of an external detector but rather originates from our designed algorithm.
Specifically, we utilize the accuracy metric (Acc) to conduct the evaluation. 
This is because under Acc measurement, we are provided by the precise location of each proposal, and only need to distinguish the target object from them. In this way, the experimental results are independent of the external detector and can reflect the ability of proposed method.
Comparing the performance of the fully supervised baseline method ReferIt3D with our weakly supervised method on Sr3D and Nr3D datasets, ReferIt3D\cite{achlioptas2020referit3d} achieves accuracy scores of 37.2 and 35.6, while our weakly supervised approach achieves scores of 36.5 and 33.8, respectively. 
These slight variations highlight the robustness of our training strategy in effectively aligning cross-modality feature representations and accurately identifying the queried object.

\textbf{Discussion on extracted target category.}
We aim to verify that the potential inaccuracies in the extracted target category, obtained through the Scene Graph Parser~\cite{zellers2018neural} and Spacy~\cite{honnibal2015improved} in Sec.~\ref{sec:3.3}, do not affect the final grounding performance.
By comparing the extracted target category with the target category provided in the dataset, we found that the extracted target category achieves an accuracy of 93\%.
Furthermore, we compare our grounding approach with the one where the target category ground truth is used, while keeping all other factors constant. In this evaluation, our method achieves an accuracy of 23.68 on Nr3D, while altering to ground truth category achieves 23.82, measured with Acc@.25.
Since the accuracy of the extracted target category does not strictly limit the performance of weakly supervised grounding, the model can tolerate a certain degree of inaccuracy in the extracted target category.
  
 

\textbf{Visualizations.}
\label{sec:1}
In Fig.~\ref{fig:vis_more}, we visualize the inference results of the category-level branch and instance-level branch to examine the effectiveness of each branch.
In Fig.\ref{fig:vis_more} (a), we show cases where the category-level branch achieves more accurate predictions than the instance-level branch. For example, in the top-left example of Fig.\ref{fig:vis_more}, the instance-level branch, without considering category-level semantic information, mistakenly identifies ``window'' as the target object, even though it does not belong to the correct category ``picture''. In contrast, the category-level branch successfully recognizes the correct category and accurately locates the desired instance, highlighting the importance of category-level differentiation.
Meanwhile, as shown in Fig.\ref{fig:vis_more} (b), when multiple distracting instances (objects of the same category) exist, the instance-level branch is more effective at distinguishing the target object from similar-looking distractors, while the category-level branch is more prone to confusion. For example, in the top-right example of Fig.\ref{fig:vis_more}, both branches identify ``backpack'', but only the instance-level branch correctly selects the target instance, whereas the category-level branch is misled by other similar objects of the same category, leading to incorrect grounding.
From these observations, we conclude that both branches have their own strengths, and by effectively integrating them, we achieve a reliable 3D weakly-supervised visual grounding.

\section{CONCLUSION AND LIMITATION}
We propose a weakly-supervised 3D visual grounding approach that effectively differentiates fine-grained categories and distinguishes instances within the same category. Our method achieves competitive performance across three benchmark datasets.
However, a limitation of our approach is the reliance on a robust object detector that can effectively recognize a wide variety of object classes, which may restrict its generalization to unseen categories. Nevertheless, with the continuous advancement of object detection techniques, we believe that in the future, object detectors will be able to achieve stable performance in an open-world setting.
\section*{ACKNOWLEDGE}
This work was supported by the grants from the National Natural Science Foundation of China 62372014 and Beijing Natural Science Foundation 4252040.

{
\bibliographystyle{IEEEtran}
\bibliography{IEEEabrv,reference}
}

\end{document}